# Adaptive Foreground and Shadow Detection in Image Sequences

**Yang Wang**
Laboratories for Information Technology
Singapore 119613
ywang@lit.org.sg

**Tele Tan**
Laboratories for Information Technology
Singapore 119613
teletan@lit.org.sg

## Abstract

This paper presents a novel method of foreground segmentation that distinguishes moving objects from their moving cast shadows in monocular image sequences. The models of background, edge information, and shadow are set up and adaptively updated. A Bayesian belief network is proposed to describe the relationships among the segmentation label, background, intensity, and edge information. The notion of Markov random field is used to encourage the spatial connectivity of the segmented regions. The solution is obtained by maximizing the posterior possibility density of the segmentation field.

## 1 INTRODUCTION

Detecting dynamic objects in image sequences is very important in such areas as surveillance and object-based coding. Accurate and efficient background removal is critical in these systems. Background subtraction based on intensity or color is a commonly used technique to identify foreground elements. The background model is built from the data and objects are segmented if they appear significantly different from the background.

To deal with illumination or object changes in the background, many researchers (Seki et al., 2000; Haritaoglu et al., 2000) have abandoned nonadaptive methods of backgrounding. The accumulation of errors in the background over time makes the method useful only in tracking applications without significant changes in the scene. Friedman and Russell (1997) classify each pixel by a probabilistic model of how that pixel looks when it is part of different classes and use an incremental EM algorithm to learn the pixel model. Stauffer and Grimson (2000) model each pixel as a mixture of Gaussians and update the model in an adaptive way. The Gaussian distributions are then evaluated to determine which are most likely to result from a background process.

Besides the nonstationariness of the background, camouflage and shadow are two classic problems of subtraction. If regions of the foreground have similar colors as the background, they can be erroneously removed. Also, shadows cast on the background can be erroneously labeled as foreground. When range data are available, depth computation from stereo cameras can be used to handle these two problems (Gordon et al., 1999). For monocular color video sequences, false segmentation caused by shadows can be minimized by computing differences in a color space that is less sensitive to intensity change (Wren et al., 1997; McKenna et al., 2000). Moreover, edge information can be utilized to improve the quality and reliability of the results (Jabri et al., 2000). Strauder et al. (1999) assumes that static edges caused by the background texture remain in regions covered by shadows and that penumbras exist at the boundary of shadows. However, this is sometimes not true due to the properties of the imaging process (Mikic et al., 2000). Mikic et al. instead approximate the change of the camera response for the shadowed region by a diagonal matrix.

On the other hand, graphical probabilistic models provide a natural tool for dealing with uncertainty and complexity through the marriage between probability theory and graph theory. In particular, Bayesian belief networks and Markov random fields are playing increasingly important roles in the design and analysis of machine learning systems (Flach, 2001). Graphical models have attracted more and more attention in vision applications such as traffic scene analysis (Koller et al., 1994), layer extraction from image sequences (Patras et al., 2001), and human motion tracking (Dockstader and Tekalp, 2001).

To solve the above mentioned problems, a unified framework of foreground segmentation for monocular intensity sequences is proposed in this paper. We introduce a Bayesian network to combine the background, intensity, and edge information. A generalized model is built for the appearance change under shadow. Camouflage is decreased by encouraging the formation of continuous segmentation regions. Parameters in the models can be updated adaptively. The solution is obtained by maximizing the posterior probability density of the segmentation field using a noniterative algorithm. Experiment shows that our method greatly improves the accuracy of segmentation.

## 2   MODEL REPRESENTATION

Given the image sequence, what we would like to do is to classify each pixel of each image as foreground (moving object), shadow, or background. The segmentation label for a point is defined as

$$s_k(\mathbf{x}) = \begin{cases} 1, \text{ if site } \mathbf{x} \text{ is in the background} \\ 2, \text{ if site } \mathbf{x} \text{ is shadowed by the foreground} \\ 3, \text{ if site } \mathbf{x} \text{ is in the foreground} \end{cases}$$

$$\forall \mathbf{x} \in \mathbf{X}, k = 1, 2, ...,$$

where $s_k(\mathbf{x})$ is the label of a single pixel $\mathbf{x}$ within the image at time $k$, and $\mathbf{X}$ is the spatial domain of the video scene. Static shadows are considered to be part of the background. The entire segmentation field is expressed compactly as $s_k$.

### 2.1   BACKGROUND MODEL

In order to segment the foreground regions in a video sequence, the system must first model the background of the video scene. Each pixel of an image acquired by the camera contains noise components. Assume that independent Gaussian noise corrupts each pixel in the scene, so that the observation model for the background becomes

$$b_k(\mathbf{x}) = \mu_{b,k}(\mathbf{x}) + n_k(\mathbf{x}), \tag{1}$$

where random variable $b_k(\mathbf{x})$ is the intensity of a single pixel $\mathbf{x}$ within the background at time $k$, and $\mu_{b,k}(\mathbf{x})$ is the intensity mean. $n_k(\mathbf{x})$ is the independent zero-mean additive noise with variance $\sigma_{b,k}^2(\mathbf{x})$ at time $k$. The parameter vector $(\mu_{b,k}(\mathbf{x}), \sigma_{b,k}^2(\mathbf{x}))^T$ is denoted as $\theta_{b,k}(\mathbf{x})$, and the entire background is expressed as $\theta_{b,k}$. For each site $\mathbf{x}$ in the background, the mean intensity and variance at time $k$ could be estimated from its history.

### 2.2   EDGE MODEL

The edge model is built by applying the edge operator to the image. This yields a horizontal difference image and a vertical difference image. For the $k$th frame $g_k$, $\mathbf{e}_{g,k}(\mathbf{x})$ is the edge vector at site $\mathbf{x} = (x_1, x_2)$,

$$\mathbf{e}_{g,k}(\mathbf{x}) = (e_{g,k}^h(\mathbf{x}), e_{g,k}^v(\mathbf{x}))^T, \tag{2a}$$

$$e_{g,k}^h(\mathbf{x}) = g_k(x_1 + 1, x_2) - g_k(x_1 - 1, x_2), \tag{2b}$$

$$e_{g,k}^v(\mathbf{x}) = g_k(x_1, x_2 + 1) - g_k(x_1, x_2 - 1), \tag{2c}$$

where $g_k(\mathbf{x})$ is the intensity of a single point $\mathbf{x}$ within the $k$th video frame, $e_{g,k}^h(\mathbf{x})$ and $e_{g,k}^v(\mathbf{x})$ are the horizontal difference and vertical difference, respectively.

Similarly, we can define the edge information for the background,

$$\mathbf{e}_{b,k}(\mathbf{x}) = (e_{b,k}^h(\mathbf{x}), e_{b,k}^v(\mathbf{x}))^T, \tag{3a}$$

$$e_{b,k}^h(\mathbf{x}) = b_k(x_1 + 1, x_2) - b_k(x_1 - 1, x_2), \tag{3b}$$

$$e_{b,k}^v(\mathbf{x}) = b_k(x_1, x_2 + 1) - b_k(x_1, x_2 - 1). \tag{3c}$$

From the background model we know that $\mathbf{e}_{b,k}(\mathbf{x})$ is of bivariate normal distribution with mean difference $\mu_{e,k}(\mathbf{x})$ and covariance matrix $\Sigma_{e,k}(\mathbf{x})$ for each site $\mathbf{x}$. $\mu_{e,k}(\mathbf{x})$ is determined by the intensity means of the four neighboring points,

$$E[e_{b,k}^h(\mathbf{x})] = \mu_{b,k}(x_1 + 1, x_2) - \mu_{b,k}(x_1 - 1, x_2), \tag{4a}$$

$$E[e_{b,k}^v(\mathbf{x})] = \mu_{b,k}(x_1, x_2 + 1) - \mu_{b,k}(x_1, x_2 - 1). \tag{4b}$$

By the independent noise assumption in the background model, $\Sigma_{e,k}(\mathbf{x})$ can be calculated from the variances of the neighboring points,

$$\text{Var}[e_{b,k}^h(\mathbf{x})] = \sigma_{b,k}^2(x_1 + 1, x_2) + \sigma_{b,k}^2(x_1 - 1, x_2), \tag{5a}$$

$$\text{Var}[e_{b,k}^v(\mathbf{x})] = \sigma_{b,k}^2(x_1, x_2 + 1) + \sigma_{b,k}^2(x_1, x_2 - 1), \tag{5b}$$

$$\text{Cov}[e_{b,k}^h(\mathbf{x}), e_{b,k}^v(\mathbf{x})] = 0. \tag{5c}$$

The parameter vector $(\mu_{e,k}(\mathbf{x}), \Sigma_{e,k}(\mathbf{x}))^T$ is denoted as $\theta_{e,k}(\mathbf{x})$, and the entire field at time $k$ is expressed as $\theta_{e,k}$. The edge model can be used to locate changes in the structure of the scenes as edges appear, disappear, or change direction.

### 2.3   SHADOW MODEL

Given the background intensity of a point $\mathbf{x}$ when illuminated, we use a linear transformation to describe the change of intensity for the same point when shadowed in the video frame at time $k$,

$$g_k(\mathbf{x}) = a_k b_k(\mathbf{x}) + c_k, \text{ if } s_k(\mathbf{x}) = 2. \tag{6}$$

When $a_k$ equals 1, the edge information will not change if the area is shadowed by the foreground. Moreover, if we extend the image input from one-channel (intensity) to multi-channel (R, G, B), the chromaticity (McKenna et al., 2000) will remain unchanged under such a linear transformation when $c_k$ is zero. So the shadow model can be viewed as the generalization of the previous assumptions. With this model for the appearance change, we can easily derive the rules for estimating means and variances for the points under shadow.

## 3 ADAPTIVE BACKGROUNDING

For static background, a sequence of background images of the scene may be recorded and the mean and variance intensity of each pixel can be calculated.

For nonstationary background, the update method is based on the ideas from Stauffer et al. (2000) and Harville et al. (2001). The recent history of each pixel, $\{g_i(\mathbf{x})\}_{1 \leq i \leq k}$, is modeled as a mixture of Gaussian distributions. The probability of the current observation is

$$p(g_k(\mathbf{x})) = \sum_{i=1}^{K} w_{i,k}(\mathbf{x}) p(g_k(\mathbf{x}) \mid \mu_{i,k}(\mathbf{x}), \sigma^2_{i,k}(\mathbf{x})), \quad (7a)$$

$$p(g_k(\mathbf{x}) \mid \mu_{i,k}(\mathbf{x}), \sigma^2_{i,k}(\mathbf{x})) = \frac{1}{\sqrt{2\pi}\sigma_{i,k}(\mathbf{x})}$$

$$\exp\{-\frac{1}{2\sigma^2_{i,k}(\mathbf{x})}[g_k(\mathbf{x}) - \mu_{i,k}(\mathbf{x})]^2\}, \quad (7b)$$

where $K$ is the number of distributions (Usually from three to five are used.), $w_{i,k}(\mathbf{x})$ is the normalized weight of the $i$th Gaussian in the mixture at time $k$, $\mu_{i,k}(\mathbf{x})$ and $\sigma^2_{i,k}(\mathbf{x})$ are the mean and variance of the $i$th Gaussian in the mixture at time $k$.

At current time $k$, each new site $g_k(\mathbf{x})$ is checked against the existing Gaussian distributions until a match (The value is within 3 standard deviations of a distribution.) is found. If the $i$th Gaussian is found to match the new observation value, parameters of the distribution are updated as follows,

$$w_{i,k}(\mathbf{x}) = (1-\alpha) w_{i,k-1}(\mathbf{x}) + \alpha, \quad (8a)$$
$$\mu_{i,k}(\mathbf{x}) = (1-\alpha) \mu_{i,k-1}(\mathbf{x}) + \alpha g_k(\mathbf{x}), \quad (8b)$$
$$\sigma^2_{i,k}(\mathbf{x}) = (1-\alpha) \sigma^2_{i,k-1}(\mathbf{x}) + \alpha(g_k(\mathbf{x}) - \mu_{i,k-1}(\mathbf{x}))^2, \quad (8c)$$

where $\alpha$ is the learning rate. (8) is equivalent to the expectation with an exponential window on the past values. For unmatched distributions, the means and variances remain the same, while the weights should be renormalized. If none of the distributions match the current pixel value, the distribution of the lowest weight is replaced with a distribution with the current value as its mean value, initially low weight, and high variance.

As the parameters of the mixture model change, the Gaussian distribution that has the most supporting evidence and the least variance is chosen as the background model for each site.

$$\theta_{b,k}(\mathbf{x}) = (\mu_{m_x,k}(\mathbf{x}), \sigma^2_{m_x,k}(\mathbf{x}))^T, \quad (9)$$

where $m_\mathbf{x} = \arg\max_i \frac{w_{i,k}(\mathbf{x})}{\sigma_{i,k}(\mathbf{x})}$. Each time after background updating, the background edge information $\theta_{e,k}$ at time $k$ can be calculated by (4) and (5).

## 4 BAYESIAN FOREGROUND DETECTION

To extract the foreground given the current frame $g_k$, difference image $e_{g,k}$, background $\theta_{b,k}$, and background edge information $\theta_{e,k}$, we wish to compute the maximum *a posteriori* (MAP) estimation of the segmentation field $s_k$. Using the Bayes' rule and ignoring the constants with respect to the unknowns,

$$\hat{s}_k = \arg\max_{s_k} p(s_k \mid \theta_{b,k}, \theta_{e,k}, g_k, e_{g,k})$$
$$= \arg\max_{s_k} p(s_k, \theta_{b,k}, \theta_{e,k}, g_k, e_{g,k})$$
$$= \arg\max_{s_k} p(\theta_{b,k}, \theta_{e,k}, g_k, e_{g,k} \mid s_k) p(s_k). \quad (10)$$

The likelihood model $p(\theta_{b,k}, \theta_{e,k}, g_k, e_{g,k} \mid s_k)$ and the prior model $p(s_k)$ must be defined for the video sequence.

### 4.1 LIKELIHOOD MODEL

Assuming conditional independence between spatially distinct observations, we factorize the likelihood model as

$$p(\theta_{b,k}, \theta_{e,k}, g_k, e_{g,k} \mid s_k)$$
$$= \prod_{\mathbf{x} \in \mathbf{X}} p(\theta_{b,k}(\mathbf{x}), \theta_{e,k}(\mathbf{x}), g_k(\mathbf{x}), e_{g,k}(\mathbf{x}) \mid s_k(\mathbf{x})). \quad (11)$$

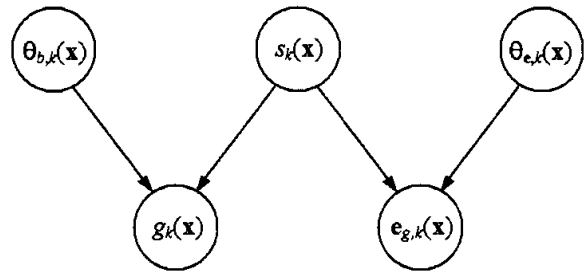

Figure 1: A Bayesian network for foreground segmentation.

The relationships among $s_k(\mathbf{x})$, $\theta_{b,k}(\mathbf{x})$, $\theta_{e,k}(\mathbf{x})$, $g_k(\mathbf{x})$, and $e_{g,k}(\mathbf{x})$ can be modeled by a Bayesian network in Figure 1. Given the segmentation label, background, and background edge information at the site, we assume that the image intensity is independent on the image edge. The conditional independence relationships implied by the belief network allow us to represent the joint more compactly (Jensen, 2001). Using the chain rule, the likelihood can be factorized as the product of the intensity likelihood $p(g_k(\mathbf{x}) \mid \theta_{b,k}(\mathbf{x}), s_k(\mathbf{x}))$ and edge likelihood $p(\mathbf{e}_{g,k}(\mathbf{x}) \mid \theta_{e,k}(\mathbf{x}), s_k(\mathbf{x}))$ at site $\mathbf{x}$.

$$\begin{aligned} &p(\theta_{b,k}(\mathbf{x}), \theta_{e,k}(\mathbf{x}), g_k(\mathbf{x}), \mathbf{e}_{g,k}(\mathbf{x}) \mid s_k(\mathbf{x})) \\ &= p(\theta_{b,k}(\mathbf{x})) p(\theta_{e,k}(\mathbf{x})) p(g_k(\mathbf{x}) \mid \theta_{b,k}(\mathbf{x}), s_k(\mathbf{x})) \\ &\quad p(\mathbf{e}_{g,k}(\mathbf{x}) \mid \theta_{e,k}(\mathbf{x}), s_k(\mathbf{x})) \\ &\propto p(g_k(\mathbf{x}) \mid \theta_{b,k}(\mathbf{x}), s_k(\mathbf{x})) p(\mathbf{e}_{g,k}(\mathbf{x}) \mid \theta_{e,k}(\mathbf{x}), s_k(\mathbf{x})). \end{aligned} \quad (12)$$

When site $\mathbf{x}$ is labeled as the background, we can calculate the intensity likelihood model $p(g_k(\mathbf{x}) \mid \theta_{b,k}(\mathbf{x}), s_k(\mathbf{x}))$ using the background model,

$$\begin{aligned} &p(g_k(\mathbf{x}) \mid \theta_{b,k}(\mathbf{x}), s_k(\mathbf{x}) = 1) \\ &= \frac{1}{\sqrt{2\pi}\sigma_{b,k}(\mathbf{x})} \exp\{-\frac{1}{2\sigma_{b,k}^2(\mathbf{x})}[g_k(\mathbf{x}) - \mu_{b,k}(\mathbf{x})]^2\}. \end{aligned} \quad (13)$$

When site $\mathbf{x}$ is shadowed, the density can be calculated by the shadow model,

$$\begin{aligned} &p(g_k(\mathbf{x}) \mid \theta_{b,k}(\mathbf{x}), s_k(\mathbf{x}) = 2) \\ &= \frac{1}{\sqrt{2\pi}a_k\sigma_{b,k}(\mathbf{x})} \exp\{-\frac{1}{2a_k^2\sigma_{b,k}^2(\mathbf{x})} \\ &\quad [g_k(\mathbf{x}) - a_k\mu_{b,k}(\mathbf{x}) - c_k]^2\}. \end{aligned} \quad (14)$$

When site $\mathbf{x}$ is labeled as the foreground, the background has no contribution to the image intensity information. Uniform distribution is assumed for the pixel. The conditional probability density becomes

$$\begin{aligned} &p(g_k(\mathbf{x}) \mid \theta_{b,k}(\mathbf{x}), s_k(\mathbf{x}) = 3) \\ &= p(g_k(\mathbf{x}) \mid s_k(\mathbf{x}) = 3) \\ &= \frac{1}{y_{max}}. \end{aligned} \quad (15)$$

Here $[0, y_{max}]$ is the intensity range for every point $\mathbf{x}$ in the scene.

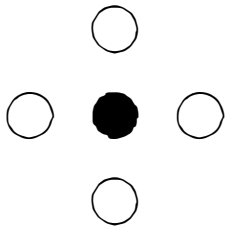

Figure 2: The first-order neighborhood system.

For each point $\mathbf{x}$, denote the set of its four nearest neighboring points by $M_\mathbf{x}$ (the first-order neighborhood, see Figure 2). Consider the spatial connectivity of the image, we assume the neighboring points have the same segmentation labels. Thus the edge likelihood $p(\mathbf{e}_{g,k}(\mathbf{x}) \mid \mathbf{e}_{b,k}(\mathbf{x}), s_k(\mathbf{x}))$ can be approximated by

$$\begin{aligned} &p(\mathbf{e}_{g,k}(\mathbf{x}) \mid \theta_{e,k}(\mathbf{x}), s_k(\mathbf{x})) \\ &\approx p(\mathbf{e}_{g,k}(\mathbf{x}) \mid \theta_{e,k}(\mathbf{x}), s_k(\mathbf{y}) = s_k(\mathbf{x}), \forall \mathbf{y} \in M_\mathbf{x}) \end{aligned}$$

$$= p(\mathbf{e}_{g,k}(\mathbf{x}) \mid \theta_{e,k}(\mathbf{x}), \prod_{\mathbf{y} \in M_\mathbf{x}} s_k(\mathbf{y}) = s_k(\mathbf{x})^{|M_\mathbf{x}|}), \quad (16)$$

where $|M_\mathbf{x}|$ is the number of elements in the set.

Similarly, when the neighborhood area $M_\mathbf{x}$ belongs to the background, the density can be computed by the edge model,

$$\begin{aligned} &p(\mathbf{e}_{g,k}(\mathbf{x}) \mid \theta_{e,k}(\mathbf{x}), \prod_{\mathbf{y} \in M_\mathbf{x}} s_k(\mathbf{y}) = 1) \\ &= \frac{1}{2\pi\sqrt{|\Sigma_{e,k}(\mathbf{x})|}} \exp\{-\frac{1}{2}[\mathbf{e}_{g,k}(\mathbf{x}) - \mu_{e,k}(\mathbf{x})]^T \Sigma_{e,k}^{-1}(\mathbf{x}) \\ &\quad [\mathbf{e}_{g,k}(\mathbf{x}) - \mu_{e,k}(\mathbf{x})]\}. \end{aligned} \quad (17)$$

When the neighborhood area $M_\mathbf{x}$ is shadowed, the density can be computed from the shadow model,

$$\begin{aligned} &p(\mathbf{e}_{g,k}(\mathbf{x}) \mid \theta_{e,k}(\mathbf{x}), \prod_{\mathbf{y} \in M_\mathbf{x}} s_k(\mathbf{y}) = 2^{|M_\mathbf{x}|}) \\ &= \frac{1}{2\pi\sqrt{|a_k^2 \Sigma_{e,k}(\mathbf{x})|}} \exp\{-\frac{1}{2a_k^2}[\mathbf{e}_{g,k}(\mathbf{x}) - a_k\mu_{e,k}(\mathbf{x})]^T \\ &\quad \Sigma_{e,k}^{-1}(\mathbf{x})[\mathbf{e}_{g,k}(\mathbf{x}) - a_k\mu_{e,k}(\mathbf{x})]\}. \end{aligned} \quad (18)$$

When neighborhood area $M_\mathbf{x}$ belongs to the foreground, we assume that the points within $M_\mathbf{x}$ are independent and identically distributed (i. i. d.). From (15), we know

$$\begin{aligned} &p(\mathbf{e}_{g,k}(\mathbf{x}) \mid \theta_{e,k}(\mathbf{x}), \prod_{\mathbf{y} \in M_\mathbf{x}} s_k(\mathbf{y}) = 3^{|M_\mathbf{x}|}) \\ &= p(\mathbf{e}_{g,k}(\mathbf{x}) \mid \prod_{\mathbf{y} \in M_\mathbf{x}} s_k(\mathbf{y}) = 3^{|M_\mathbf{x}|}) \\ &= p(e_{g,k}^h(\mathbf{x}) \mid \prod_{\mathbf{y} \in M_\mathbf{x}} s_k(\mathbf{y}) = 3^{|M_\mathbf{x}|}) \\ &\quad p(e_{g,k}^v(\mathbf{x}) \mid \prod_{\mathbf{y} \in M_\mathbf{x}} s_k(\mathbf{y}) = 3^{|M_\mathbf{x}|}) \\ &= (\frac{1}{y_{max}} - \frac{|e_{g,k}^h(\mathbf{x})|}{y_{max}^2})(\frac{1}{y_{max}} - \frac{|e_{g,k}^v(\mathbf{x})|}{y_{max}^2}). \end{aligned} \quad (19)$$

### 4.2 PRIOR MODEL

The prior model $p(s_k)$ represents the prior probability of the segmentation field. We model the density by a Markov random field (Geman and Geman, 1984). That is, if $N_\mathbf{x}$ is a neighborhood of the pixel at $\mathbf{x}$, then the conditional distribution of a single variable at $\mathbf{x}$ is completely specified by the variables within its neighborhood $N_\mathbf{x}$. According to the Hammersley-Clifford theorem, the density is given by a Gibbs density that has the following form (Tekalp, 1995):

$$p(s_k) \propto \exp\{-\sum_{c \in C} V_k(s_k(\mathbf{x}) \mid \mathbf{x} \in c)\}, \quad (20)$$

where $C$ is the set of all cliques $c$, and $V_k$ is the clique potential function at time $k$. A clique is a set of points that are neighbors of each other. The clique potential depends only on the pixels that belong to clique $c$. Only one-pixel and two-pixel cliques are used in our work.

The single-pixel clique potentials can be defined as

$$V_{1,k}(s_k(\mathbf{x})) = \eta_{s_k(\mathbf{x}),k} \,. \qquad (21)$$

They reflect our prior knowledge of the probabilities of different region types. The lower the value of $\eta_{s_k(\mathbf{x}),k}$, the more likely that the point $\mathbf{x}$ is labeled as $s_k(\mathbf{x})$ at time $k$.

Spatial connectivity can be imposed by the following two-pixel clique potential,

$$V_2(s_k(\mathbf{x}), s_k(\mathbf{y})) = \frac{1}{\|\mathbf{x}-\mathbf{y}\|^2} (1 - \delta(s_k(\mathbf{x}) - s_k(\mathbf{y}))), (22)$$

where $\delta(\cdot)$ is the Dirac delta function, and $\|\cdot\|$ denotes the Euclidian distance. Thus two neighboring pixels are more likely to belong to the same class than to different classes. The constraint becomes stronger with decrease of the distance between the neighboring sites.

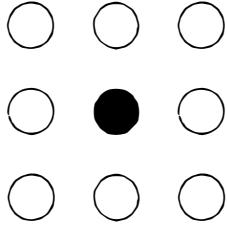

Figure 3: The second-order neighborhood system.

Combining the above models, the Bayesian MAP estimate is obtained by minimizing the objective function

$$F_k(s_k) = \sum_{\mathbf{x} \in X} U_{1,k}(\mathbf{x}, s_k(\mathbf{x})) + \sum_{\mathbf{x} \in X} U_{2,k}(\mathbf{x}, s_k(\mathbf{x})) +$$
$$\lambda_1 \sum_{\mathbf{x} \in X} V_{1,k}(s_k(\mathbf{x})) + \lambda_2 \sum_{\{\mathbf{x},\mathbf{y}\} \in C} V_2(s_k(\mathbf{x}), s_k(\mathbf{y})),$$
$$(23)$$

where $U_{1,k}(\mathbf{x}, s_k(\mathbf{x})) = -\ln p(g_k(\mathbf{x}) | \theta_{b,k}(\mathbf{x}), s_k(\mathbf{x}))$, and $U_{2,k}(\mathbf{x}, s_k(\mathbf{x})) = -\ln p(e_{g,k}(\mathbf{x}) | \theta_{e,k}(\mathbf{x}), s_k(\mathbf{x}))$. The parameters $\eta_{1,k}, \eta_{2,k}, \eta_{3,k}, \lambda_1$ and $\lambda_2$ should be determined carefully to control the affection of each term in (23).

## 5 IMPLEMENTATION

### 5.1 PARAMETER DETERMINATION

After the segmentation of the $k$th frame, denote the set of points labeled as $s$ ($s = 1, 2, 3$) by $\mathbf{X}_{s,k}$. The single-pixel clique potential can be reestimated as

$$\eta^*_{i,k} = -\frac{|\mathbf{X}_{i,k}|}{\sum_s |\mathbf{X}_{s,k}|}, i = 1, 2, 3. \qquad (24)$$

With the learning rate $\alpha$, $\eta_{i,k+1}$ can be updated in an adaptive way.

$$\eta_{i,k+1} = (1-\alpha)\eta_{i,k} + \alpha \eta^*_{i,k} \,. \qquad (25)$$

The parameters of the linear transformation in the shadow model can be reestimated from the set $\{(g_k(\mathbf{x}), b_k(\mathbf{x})) | \mathbf{x} \in \mathbf{X}_{2,k}\}$ by the least squares method,

$$a^*_k = \frac{\sum_{\mathbf{x} \in \mathbf{X}_{2,k}} g_k(\mathbf{x}) \sum_{\mathbf{x} \in \mathbf{X}_{2,k}} b_k(\mathbf{x}) - |\mathbf{X}_{2,k}| \sum_{\mathbf{x} \in \mathbf{X}_{2,k}} g_k(\mathbf{x}) b_k(\mathbf{x})}{(\sum_{\mathbf{x} \in \mathbf{X}_{2,k}} b_k(\mathbf{x}))^2 - |\mathbf{X}_{2,k}| \sum_{\mathbf{x} \in \mathbf{X}_{2,k}} b_k^2(\mathbf{x})},$$
$$(26a)$$

$$c^*_k = \frac{\sum_{\mathbf{x} \in \mathbf{X}_{2,k}} g_k(\mathbf{x}) - a^*_k \sum_{\mathbf{x} \in \mathbf{X}_{2,k}} b_k(\mathbf{x})}{|\mathbf{X}_{2,k}|}. \qquad (26b)$$

The shadow model is then updated adaptively.

$$a_{k+1} = (1 + \eta^*_{2,k}\alpha)a_k - \eta^*_{2,k}\alpha a^*_k, \qquad (27a)$$

$$c_{k+1} = (1 + \eta^*_{2,k}\alpha)c_k - \eta^*_{2,k}\alpha c^*_k. \qquad (27b)$$

In (27) the effective learning rate $-\eta^*_{2,k}\alpha$ changes with the ratio of shadowed points in the scene. This helps make a robust updating process especially for the frames where there are only few points shadowed.

The initial values are set as $a_1 = 0.5$, $c_1 = 0$, and $\eta_{1,1} = \eta_{2,1} = \eta_{3,1} = -\frac{1}{3}$. Parameters $\alpha$, $\lambda_1$, and $\lambda_2$ reflect the importance of previous knowledge, one-pixel clique potential, and two-pixel clique potential, respectively. They are determined manually in this paper.

### 5.2 OPTIMIZATION

Obviously, there is no simple method of performing the optimization in (23), furthermore, the objective function does not have a unique minimum since it is nonconvex in terms of $s_k(\mathbf{x})$. To arrive at a sub-optimal estimate, we use a local technique known as highest confidence first (HCF). HCF is a non-iterative, deterministic algorithm for combinational minimization (Chou and Brown, 1990). It is guaranteed to reach a local minimum of the objective function after a finite number of steps. Its feature is the introduction of a special uncommitted label and the strategy for "committing" a site (Li, 1995). Denote the uncommitted label by 0, the original label set is augmented by this label into $\{0,1,2,3\}$.

Given the labels of the points within the neighborhood $N_x$, the conditional posterior potential for a point **x** at time $k$ is defined as

$$f_k(\mathbf{x}, s_k(\mathbf{x})) = U_{1,k}(\mathbf{x}, s_k(\mathbf{x})) + U_{2,k}(\mathbf{x}, s_k(\mathbf{x}))$$
$$+ \lambda_1 V_{1,k}(s_k(\mathbf{x})) + \lambda_2 \sum_{y \in N_x} V_2^*(s_k(\mathbf{x}), s_k(\mathbf{y})), \quad (28a)$$

$$V_2^*(s_k(\mathbf{x}), s_k(\mathbf{y})) = \begin{cases} 0, & \text{if } s_k(\mathbf{y}) = 0. \\ V_2(s_k(\mathbf{x}), s_k(\mathbf{y})), & \text{otherwize.} \end{cases}$$
(28b)

In our work, the second-order neighborhood system is used (see Figure 3). Based on the conditional posterior potential, we can define the "stability" measure of site **x**.

$$S_k(\mathbf{x}, s_k(\mathbf{x})) =$$
$$\begin{cases} -\min_{s \neq 0, s_{\min,k}(\mathbf{x})} [f_k(\mathbf{x}, s) - f_k(\mathbf{x}, s_{\min,k}(\mathbf{x}))], & \text{if } s_k(\mathbf{x}) = 0. \\ \min_{s \neq 0, s_k(\mathbf{x})} [f_k(\mathbf{x}, s) - f_k(\mathbf{x}, s_k(\mathbf{x}))], & \text{otherwise.} \end{cases}$$
(29)

where $s_{\min,k}(\mathbf{x}) = \arg \min_{s \neq 0} f_k(\mathbf{x}, s)$.

The "stability" measure $S_k(\mathbf{x}, s_k(\mathbf{x}))$ is used to determine the order in which the sites are to be visited. Initially, all points are labeled uncommitted. Once a non-zero label is assigned to an uncommitted site, the site is committed and cannot return to the uncommitted state. However, the label of a committed site can be changed to another non-zero value. The optimization procedure terminates when the objective function (23) can no longer be decreased by reassignment of the labels.

## 6 EXPERIMENTAL RESULTS AND DISCUSSION

The algorithm has been tested by indoor and outdoor sequences. To reduce the heavy computation afford, we assume $\sigma_{b,k}^2(\mathbf{x}) = \sigma_{b,k}^2$ for every point at the step of Bayesian foreground detection. Figure 4 shows the segmentation results for the "aerobic" sequence. Figure 4a is one frame of the sequence, and Figure 4b is the estimated background. The segmentation results of both simple background subtraction and our method are shown in Figure 4c-4f. Comparing with the results of simple background subtraction, the accuracy of the calculated object location is greatly improved by the proposed approach. The moving cast shadows (the gray regions in Figure 4d) are exactly removed from the foreground. The flickering background pixels that simple background subtraction will detect as foreground are correctly classified by our algorithm. The camouflage at the neck makes the head almost separated from the body in figure 4e, while the disadvantage is successfully overcome in figure 4f.

We can see that the boundary pixels of the background are classified as foreground by our method. This is caused by the change of the edge information at these points. To solve this problem, a thinning algorithm could be performed to remove the boundary points.

During the segmentation process in section 4, the density of the image intensity at site **x** is modeled as

$$p(g_k(\mathbf{x}) \mid \theta_{b,k}(\mathbf{x}))$$
$$= \sum_{s_k(\mathbf{x})=1}^{3} p(s_k(\mathbf{x})) p(g_k(\mathbf{x}) \mid \theta_{b,k}(\mathbf{x}), s_k(\mathbf{x})). \quad (30)$$

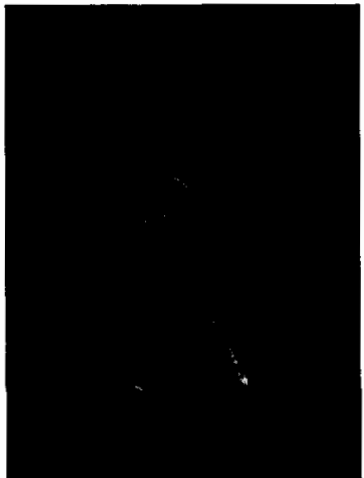 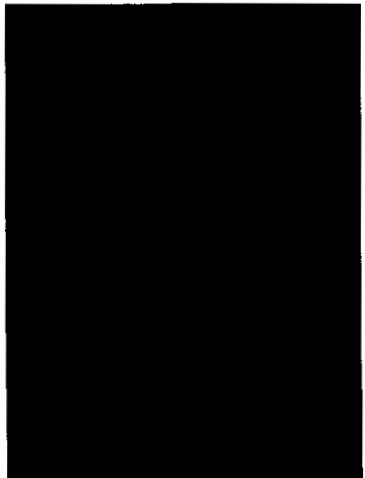 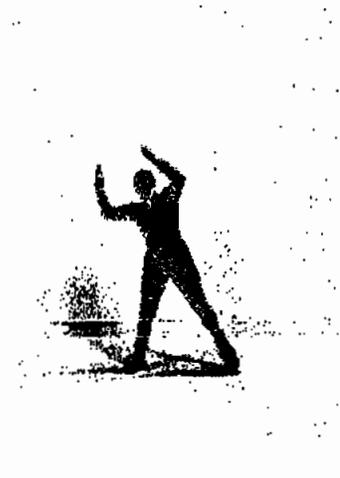

(a)                 (b)                 (c)

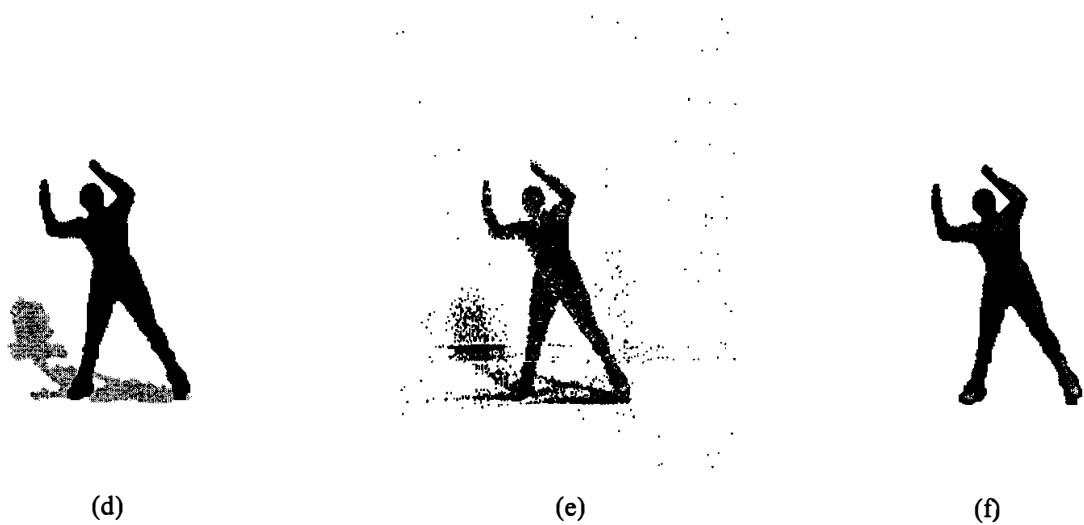

Figure 4: (a) One frame of the "aerobic" sequence. (b) The estimated background. (c) The segmentation result of simple background subtraction. (d) The segmentation result of the proposed algorithm. (e) The foreground detected by simple background subtraction. (f) The foreground detected by the proposed algorithm.

Comparing with the right side of (7a) in the case of $K = 3$, it can be found that uniform distribution is assumed for the foreground in (30), while Gaussian distribution is assumed in (7a). (30) could be thought as the improvement of (7a), since in the foreground there is no particular reason to prefer one value over any other. However, the mixture different kind of distributions is much harder to estimate than the mixture of only Gaussians. Since foreground regions usually have large variances, from (9) we can see that such a difference will not affect the backgrounding results.

## 7 CONCLUSION

In this paper we have presented an adaptive approach for foreground segmentation and shadow detection in monocular image sequences. Graphical probabilistic models are employed in our method. In our work, we could identify three sources of information that can help in detecting objects and shadows. The first is edge information, the difference images help locate changes in the scene. The second source of information is spatial, objects and shadows usually form continuous regions, and the third is temporal, the models are updated from previous segmentation results.

Experimental results show that our method successfully deals with nonstationary background, camouflage and shadows in the video sequence. Moreover, the algorithm can be easily implemented for color image sequences. How to decrease the computation load of the optimization process and automatically determine all the parameters in our model could be our future study.